\RequirePackage{fix-cm}
\documentclass[smallextended]{svjour3}       
\smartqed  
\usepackage{graphicx}
\usepackage{tabularx}
\usepackage{colortbl}
\usepackage{subfigure}
\usepackage{color}
\begin{document}

\title{A Benchmark for Endoluminal Scene Segmentation of Colonoscopy Images
}
\titlerunning{EndoScene benchmark}        

\author{David V\'azquez$^{1,2}$ \and Jorge Bernal$^{1}$ \and F. Javier S\'anchez$^{1}$ \and Gloria Fern\'andez-Esparrach$^{4}$ \and Antonio M. L\'opez$^{1,2}$ \and Adriana Romero$^{2}$\and Michal Drozdzal$^{3, 5}$ \and Aaron Courville$^{2}$
}
\authorrunning{David V\'azquez et al.}

\institute{David V\'azquez \at \email{dvazquez@cvc.uab.es}\\
$^{1}$ Computer Vision Center, Computer Science Department, Universitat Autonoma de Barcelona, Spain\\
$^{2}$ Montreal Institute for Learning Algorithms, Universit\'e de Montr\'eal, Canada\\
$^{3}$ \'{E}cole Polytechnique de Montr\'{e}al, Montr\'{e}al, Canada\\
$^{4}$ Endoscopy Unit, Gastroenterology Service, CIBERHED, IDIBAPS, Hospital Clinic, Universidad de Barcelona, Spain\\
$^{5}$ Imagia Inc., Montr\'{e}al, Canada\\
}

\date{Received: 07 Nov 2016 / Accepted: date}

\maketitle

\begin{abstract}
Colorectal cancer (CRC) is the third cause of cancer death worldwide. Currently, the standard approach to reduce CRC-related mortality is to perform regular screening in search for polyps and colonoscopy is the screening tool of choice. The main limitations of this screening procedure are polyp miss-rate and inability to perform visual assessment of polyp malignancy. These drawbacks can be reduced by designing Decision Support Systems (DSS) aiming to help clinicians in the different stages of the procedure by providing endoluminal scene segmentation. Thus, in this paper, we introduce an extended benchmark of colonoscopy image, with the hope of establishing a \emph{new strong benchmark for colonoscopy image analysis research}. We provide new baselines on this dataset by training standard fully convolutional networks (FCN) for semantic segmentation  and \emph{significantly outperforming, without any further post-processing}, prior results in endoluminal scene segmentation. 
\end{abstract}

\section{Introduction} 
\label{sec:intro} 
\begin{figure}[t!]
\centering
\subfigure[]{\includegraphics[width=0.55\textwidth]{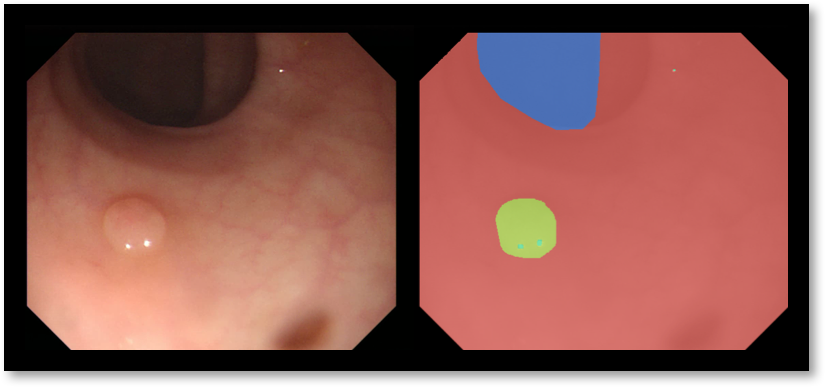}\label{fig:scene}}\hfill
\subfigure[]{\includegraphics[width=0.75\textwidth]{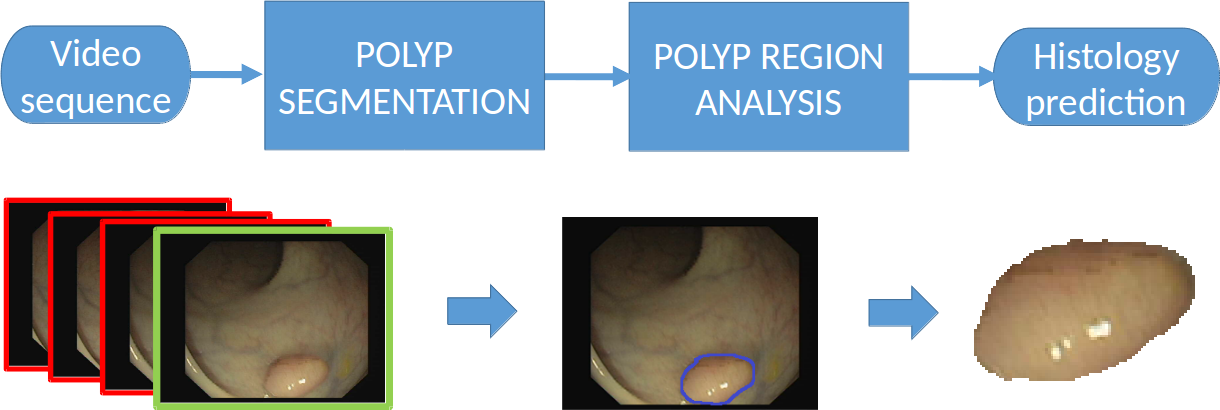}\label{fig:pipeline}}\hfill
\caption{(a) Colonoscopy image and corresponding labeling: blue for lumen, red for background (mucosa wall), and green for polyp, (b) Proposed pipeline of a decision support system for colonoscopy.}
\label{fig:first_fig}
\end{figure}

Colorectal cancer (CRC) is the third cause of cancer death worldwide \cite{cancerstats}. CRC arises from adenomatous polyps (adenomas) which are initially benign; however, with time some of them can become malignant. Currently, the standard approach to reduce CRC-related mortality is to perform regular screening in search for polyps and colonoscopy is the screening tool of choice. During the examination, clinicians visually inspect intestinal wall (see Figure \ref{fig:scene} for an example of intestinal scene) in search of polyps; once found they are resected for histological analysis.

The main limitations of colonoscopy are its associated polyp miss-rate (small/flat polyps or the ones hidden behind intestine folds can be missed \cite{leufkens2012factors}) and the fact that polyp's malignancy degree is only known after histological analysis. These drawbacks can be reduced by developing new colonoscopy modalities to improve visualization  (e.g. High Definition imaging, Narrow Band Imaging (NBI) \cite{machida2004narrow} and magnification endoscopes \cite{bruno2003magnification}) and/or by developing Decision Support Systems (DSS) aiming to help clinicians in the different stages of the procedure. Ideally, DSS should be able to detect, segment and assess the malignancy degree (e.g. by optical biopsy \cite{roy2011colonoscopic}) of polyps during colonoscopy procedure (Figure \ref{fig:pipeline} depicts steps of DSS in colonoscopy procedure). 

The development of DSS for colonoscopy has been an active research topic during the last decades. The majority of available works on optical colonoscopy are focused on polyp detection (e.g. see \cite{bernal2015wm,Silva2014,gross2009comparison,park2016colonoscopic,ribeiro2016colonic,tajbakhsh2015automated}) and only few works address the problems of endoluminal scene segmentation. 

Endoluminal scene segmentation is of crucial relevance for clinical applications \cite{bernal2014polyp,bernal2015wm,nunez2014impact,bernal2014discarding}. First, segmenting \emph{lumen} is relevant to help clinicians in navigation during the procedure. Second, segmenting \emph{polyps} is important to indicate the area covered by a potential lesion that should be carefully inspected by clinicians. Finally, \emph{specular highlight} have proven to be useful in reducing polyp detection false positive ratio in the context of hand-crafted methods \cite{bernal2015wm,bernal2013impact}. 

In the recent years, convolutional neural networks (CNNs) have become a \emph{de facto} standard in computer vision, achieving state-of-the-art performance in tasks such as image classification, object detection and semantic segmentation; and making traditional methods based on hand-crafted features obsolete. Two major components in this groundbreaking progress were the availability of increased computational power (GPUs) and the introduction of large labeled datasets \cite{imagenet,mscoco}.

Despite the additional difficulty of having limited amounts of labeled data, CNNs have successfully been applied to a variety of medical imaging tasks, by resorting to aggressive data augmentation techniques \cite{ronneberger2015u,Drozdzal16}. More precisely, CNNs have excelled at semantic segmentation tasks in medical imaging, such as the EM ISBI 2012 dataset~\cite{EM_data}, BRATS~\cite{Brats} or MS lesions~\cite{MsLesions}, where the top entries are built on CNNs \cite{ronneberger2015u,0011QCH16,HavaeiDWBCBPJL15,Brosch,Drozdzal16}. Surprisingly, to the best of our knowledge, CNNs have not been applied to semantic segmentation of colonoscopy data. We associate this to the lack of large publicly available annotated databases, which is needed in order to train and validate such networks. 

In this paper, we aim to overcome this limitation and introduce an extended benchmark of colonoscopy images by combining the two largest datasets of colonoscopy images \cite{bernal2012towards,bernal2015wm} and incorporating additional annotations to segment lumen and specular highlights, with the hope of establishing a \emph{new strong benchmark for colonoscopy image analysis research}. We provide new baselines on this dataset by training standard fully convolutional networks (FCN) for semantic segmentation \cite{long2015fully}, and \emph{significantly outperforming, without any further post-processing}, prior results in endoluminal scene segmentation. 

\vspace{0.2cm}

Therefore, the contributions of this paper are two-fold:
\vspace{-.1cm}
\begin{enumerate}
\item Extended benchmark for colonoscopy image segmentation.
\item New state-of-the-art in colonoscopy image segmentation.
\end{enumerate}

The rest of the paper is organized as follows. In Section \ref{sec:bench}, we present the new extended benchmark, including the introduction of datasets as well as the performance metrics. After that, in Section \ref{sec:deep}, we introduce the FCN architecture used as baseline for the new endoluminal scene segmentation benchmark. Then, in Section \ref{sec:results}, we show qualitative and quantitative experimental results. Finally, Section \ref{sec:conclusions} concludes the paper. 

\section{Endoluminal scene segmentation benchmark}
\label{sec:bench} 
In this section, we describe the endoluminal scene segmentation benchmark, including evaluation metrics.

\subsection{Dataset}
\label{ssec:dataset}

Inspired by already published benchmarks for polyp detection, proposed within a challenge held in conjunction with MICCAI 2015\footnote{http://endovis.grand-challenge.org}, we introduce a benchmark for endoluminal scene object segmentation. 

We combine \textbf{CVC-ColonDB} and \textbf{CVC-ClinicDB} into a new dataset (\textbf{EndoScene}) composed of 912 images obtained from 44 video sequences acquired from 36 patients. 
\begin{itemize}
\item \textbf{CVC-ColonDB} contains 300 images with associated polyp and background (here, mucosa and lumen) segmentation masks obtained from 13 polyp video sequences acquired from 13 patients.
\item \textbf{CVC-ClinicDB} contains 612 images with associated polyp and background (here, mucosa and lumen) segmentation masks obtained from 31 polyp video sequences acquired from 23 patients. 
\end{itemize}

\begin{figure}[t!]
\centering
\subfigure[]{\includegraphics[width=0.24\textwidth]{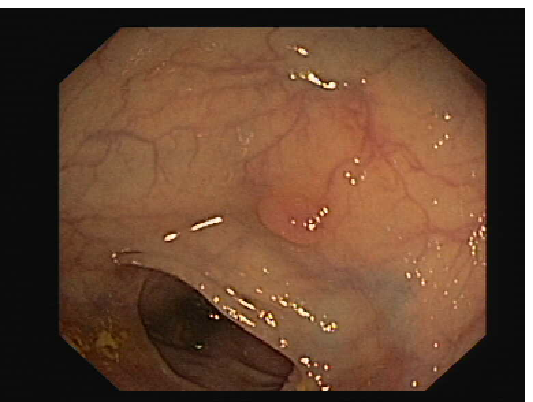}}
\subfigure[]{\includegraphics[width=0.24\textwidth]{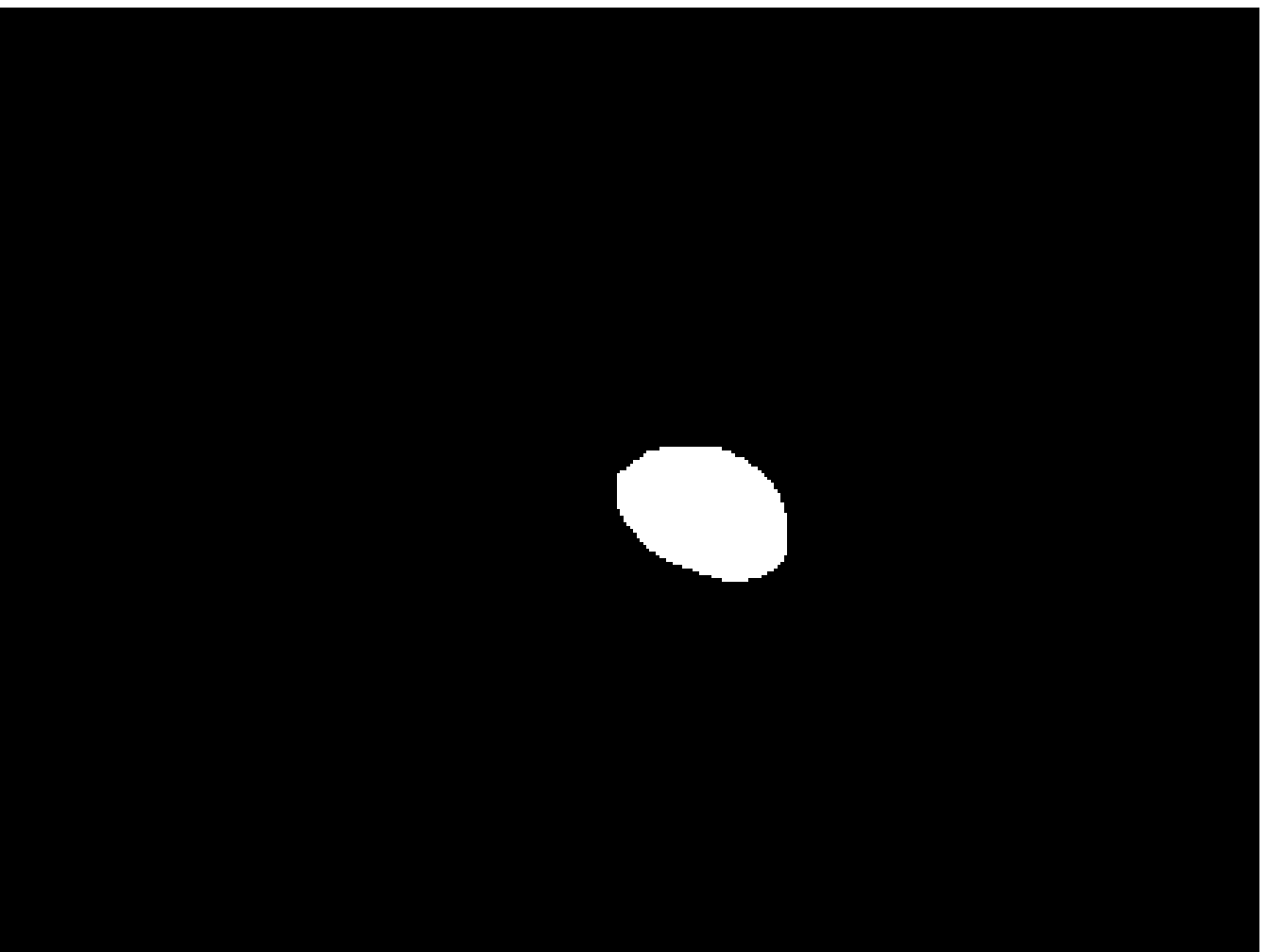}}
\subfigure[]{\includegraphics[width=0.24\textwidth]{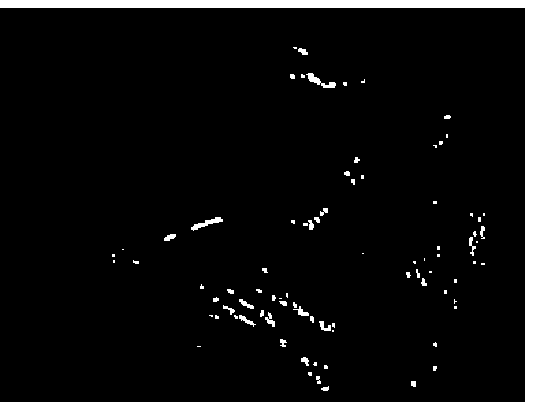}}
\subfigure[]{\includegraphics[width=0.24\textwidth]{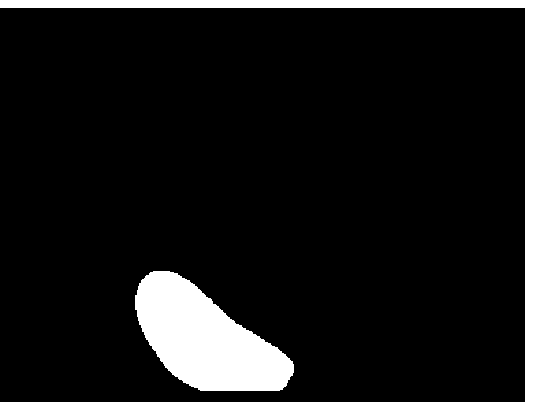}}
\caption{Example of a colonoscopy image and its corresponding ground truth: (a) Original image, (b) Polyp mask, (c) Specular highlights mask and (d) lumen mask.}
\label{fig:databases}
\end{figure}

We extend the old annotations to account for lumen, specular highlights as well as a void class for black borders present in each frame. In the new annotations, background only contains mucosa (intestinal wall). Please refer to Table \ref{tab:databases} for datasets details and to Figure \ref{fig:databases} for a dataset sample. 

We split the resulting dataset into three sets: training, validation and test containing 60\%, 20\% and 20\% images respectively. We impose the constraint that one patient can not be in different sets. As a result, the final training set contains 20 patients and 547 frames, the validation set contains 8 patients and 183 frames; and test set contains 8 patients and 182 frames. The dataset is available online\footnote{http://adas.cvc.uab.es/endoscene}.

\begin{table}[t!]
\begin{center}
\caption{Summary of prior database content. All frames show at least one polyp.}
\begin{scriptsize}
\begin{tabular}{|c||c|c|c|c|c|}
  \hline
  \textbf{Database} & \textbf{\# patients} & \textbf{\# seq.} & \textbf{\# frames} & \textbf{Resolution} & \textbf{Annotations}\\
  \hline  \hline
  \begin{tabular}{@{}c@{}}CVC- \\ ColonDB\end{tabular} & 13 & 13 & 300 & 500x574 &  \begin{tabular}{@{}c@{}}polyp \\ background\end{tabular} \\\hline
  \begin{tabular}{@{}c@{}}CVC- \\ ClinicDB\end{tabular} & 23 & 31 & 612 & 384x288 & \begin{tabular}{@{}c@{}}polyp \\ background\end{tabular} \\
  \hline
   \hline
  EndoScene & 36 & 44 & 912 & 500x574 \& 384x288 & \begin{tabular}{@{}c@{}}polyp \\ lumen \\ background \\specularity \\ border (void)\end{tabular} \\
  \hline
\end{tabular}
\end{scriptsize}
\label{tab:databases}
\end{center}
\end{table}

\subsection{Metrics}

We use Intersection over Union (IoU), also known as \emph{Jaccard index}, and per pixel accuracy as segmentation metrics. These metrics are commonly used in medical image segmentation tasks \cite{cha2016urinary,prastawa2004brain}. 

We compute the mean of per-class IoU. Each per-class IoU is computed over a validation/test set according to following formula:
\begin{equation}
IoU(\mathrm{PR}(class), \mathrm{GT}(class)) = \frac{|\mathrm{PR}(class) \bigcap \mathrm{GT}(class)|}{|\mathrm{PR}(class) \bigcup \mathrm{GT}(class)|},
\end{equation}
where $\mathrm{PR}$ represents the binary mask produced by the segmentation method, $\mathrm{GT}$ represents the ground truth mask, $\bigcap$ represents set intersection and $\bigcup$ represents set union. 

We compute the mean global accuracy for each set as follows:
\begin{equation}
Acc(\mathrm{PR}, \mathrm{GT}) = \frac{\# \mathrm{TP}}{\# \mathrm{pixels}},
\end{equation}
where $\mathrm{TP}$ represents the number of true positives. 

Notably, this new benchmark might as well be used for the relevant task of polyp localization. In that case, we follow Pascal VOC challenge metrics \cite{Everingham10} and determine that a polyp is localized if it has a high overlap degree with its associated ground truth, namely:
\begin{equation}
IoU(\mathrm{PR}(polyp), \mathrm{GT}(polyp)) > 0.5,
\end{equation}
where the metric is computed for each polyp independently, and averaged per set to give a final score. 

\section{Baseline} 
\label{sec:deep}

CNN are a standard architecture used for tasks, where a single prediction per input is expected (e.g. image classification). Such architectures capture hierarchical representations of the input data by stacking blocks of convolutional, non-linearity and pooling layers on top of each other. Convolutional layers extract local features. Non-linearity layers allow deep networks to learn non-linear mappings of the input data. Pooling layers reduce the spatial resolution of the representation maps by aggregating local statistics.

FCNs \cite{long2015fully,ronneberger2015u} were introduced in the computer vision and medical imaging communities in the context of semantic segmentation. FCNs naturally extend CNNs to tackle per pixel prediction problems, by adding upsampling layers to recover the spatial resolution of the input at the output layer. As a consequence, FCNs can process images of arbitrary size. In order to compensate for the resolution loss induced by pooling layers, FCNs introduce skip connections between their downsampling and upsampling paths. Skip connections help the upsampling path recover fine-grained information from the downsampling layers. 

We implemented FCN8 architecture from \cite{long2015fully}\footnote{https://github.com/jbernoz/deeppolyp}. and trained the network by means of stochastic gradient descent with rmsprop adaptive learning rate \cite{rmsprop}. The validation split is used to early-stop the training, we monitor mean IoU for validation set and use patience of 50. We used a mini-batch size of 10 images. The input image is normalized in the range 0-1. We randomly crop the training images to $224\times224$ pixels. As regularization, we use dropout\cite{Srivastava2014} of 0.5, as mentioned in the paper \cite{long2015fully}. We do not use any weight decay.

As described in Section \ref{ssec:dataset}, colonoscopy images have a black border that we consider as a void class. Void classes do not influence the computation of the loss nor the metrics of any set, since the pixels marked as void class are ignored. As the number of pixels per class is unbalanced, in some experiments, we apply the median frequency balancing of \cite{EigenF14}.

During training, we experiment with data augmentation techniques such as random cropping, rotations, zooming, sharing and elastic transformations.

\section{Experimental results}
\label{sec:results}

In this section, we report semantic segmentation and polyp localization results on the new benchmark. 

\subsection{Endoluminal Scene Semantic Segmentation}

In this section, we first analyze the influence of different data augmentation techniques. Second, we evaluate the effect of having different number of endoluminal classes on polyp segmentation results. Finally, we compare our results with previously published methods.

\subsubsection{Influence of data augmentation}

Table \ref{tab:results_data_aug_valid} presents an analysis on the influence of different data augmentation techniques and their impact on the validation performance. We evaluate random zoom from 0.9 to 1.1, rotations from 0 to 180 degrees, shearing from 0 to 0.4 and warping with $\sigma$ ranging from 0 to 10. Finally, we evaluate the combination of all the data augmentation techniques. 

As shown in the table, polyps significantly benefit from all data augmentation methods, in particular, from warping. Note that warping applies small elastic deformation locally, accounting for many realistic variations in the polyp shape. Rotation and zoom also have a strong positive impact on the polyp segmentation performance. It goes without saying that such transformations are the least aggressive ones, since they do not alter the polyp appearance. Shearing is most likely the most aggressive transformation, since it changes the polyp appearance and might, in some cases, result in unrealistic deformations.

While for lumen it is difficult to draw any strong conclusions, it looks like zooming and warping slightly deteriorate the performance, whereas shearing and rotation slightly improve it. As for specularity highlights, all the data augmentation techniques that we tested significantly boost the segmentation results. Finally, background (mucosa) shows only slight improvement when incorporating data augmentations. This is not surprising, given its predominance throughout the data it could be even considered background.

Overall, combining all the discussed data augmentation techniques, leads to better results in terms of mean IoU and mean global accuracy. More precisely, we increase the mean IoU by $4.51\%$ and the global mean accuracy by $1.52\%$

\begin{table}
\begin{center}
\begin{tabular}{|c|c|c|c|c||c|c|}
\hline
\begin{tabular}{@{}c@{}}\textbf{Data} \\ \textbf{Augmentation}\end{tabular}
& \begin{tabular}{@{}c@{}}\textbf{IoU} \\ \textbf{background}\end{tabular}
& \begin{tabular}{@{}c@{}}\textbf{IoU} \\ \textbf{polyp}\end{tabular}
& \begin{tabular}{@{}c@{}}\textbf{IoU} \\ \textbf{lumen}\end{tabular}
& \begin{tabular}{@{}c@{}}\textbf{IoU} \\ \textbf{spec.}\end{tabular}
& \begin{tabular}{@{}c@{}}\textbf{IoU} \\ \textbf{mean}\end{tabular}
& \begin{tabular}{@{}c@{}}\textbf{Acc} \\ \textbf{mean}\end{tabular}\\
\hline
\hline
None & 88.93 & 44.45 & 54.02 & 25.54 & 57.88 & 92.48 \\ 
\hline
Zoom   & 89.89 & 52.73 & 51.15 & 37.10 & 57.72 & 90.72  \\
\hline
Warp   & 90.00 & 54.00 & 49.69 & \textbf{37.27} & 58.97 & 90.93  \\
\hline
Shear   & 89.60 & 46.61 & 54.27 & 36.86 & 56.83 & 90.49  \\
\hline
Rotation   & 90.52 & 52.83 & \textbf{56.39} & 35.81 & 58.89 & 91.38  \\
\hline
Combination & \textbf{92.62} & \textbf{54.82} & 55.08 & 35.75 & \textbf{59.57} & \textbf{93.02} \\
\hline
\end{tabular}
\end{center}
\caption{FCN8 endoluminal scene semantic segmentation results for different data augmentation techniques. The results are reported on validation set.}
\label{tab:results_data_aug_valid}
\end{table}

\subsubsection{Influence of number of classes}

Table \ref{tab:nb_classes_valid} presents endoluminal scene semantic segmentation results for different number of classes. As shown in the table, using more underrepresented classes such as lumen or specular highlights makes the optimization problem more difficult. As expected and contrary to hand-crafted segmentation methods, when considering polyp segmentation, deep learning based approaches do not suffer from specular highlights, showing the robustness of the learnt features towards saturation zones in colonoscopy images.

Best results for polyp segmentation are obtained in the 2 classes scenario (polyp vs background). However, segmenting lumen is a relevant clinical problem as mentioned in Section \ref{sec:intro}. Results achieved in the 3 classes scenario are very encouraging, with a IoU higher than $50\%$ for both polyp and lumen classes.

\begin{table}
\begin{center}
\begin{tabular}{|c|c|c|c|c||c|c|}
\hline
\begin{tabular}{@{}c@{}}\textbf{\#} \\ \textbf{classes}\end{tabular}
& \begin{tabular}{@{}c@{}}\textbf{IoU} \\ \textbf{background}\end{tabular}
& \begin{tabular}{@{}c@{}}\textbf{IoU} \\ \textbf{polyp}\end{tabular}
& \begin{tabular}{@{}c@{}}\textbf{IoU} \\ \textbf{lumen}\end{tabular}
& \begin{tabular}{@{}c@{}}\textbf{IoU} \\ \textbf{spec.}\end{tabular}
& \begin{tabular}{@{}c@{}}\textbf{IoU} \\ \textbf{mean}\end{tabular}
& \begin{tabular}{@{}c@{}}\textbf{Acc} \\ \textbf{mean}\end{tabular}\\
\hline \hline
4& 92.07 & 39.37 & \textbf{59.55} & \textbf{40.52} &  57.88 & 92.48\\ 
\hline
3 & 92.19 & 50.70 & 56.48 &  --  & 66.46 & 92.82\\ 
\hline
2 & \textbf{96.63} & \textbf{56.07} &   --   &  --  &  \textbf{76.35} & \textbf{96.77}\\
\hline 
\end{tabular}
\end{center}
\caption{FCN8 endoluminal scene semantic segmentation results for different number of classes. The results are reported on validation set. In all cases, we selected the model that provided best validation results (with or without class balancing).}
\label{tab:nb_classes_valid}
\end{table}

\subsubsection{Comparison to state-of-the-art}
Finally, we evaluate the FCN model on the test set. We compare our results to the combination of previously published hand-crafted methods: 1) \cite{bernal2014polyp} an energy-map based method for polyp segmentation, 2) \cite{bernal2014discarding} a watershed-based method for lumen segmentation  and  3) \cite{bernal2013impact} for specular highlights segmentation.  

\begin{table}
\setlength\tabcolsep{4.5pt} 
\begin{center}
\begin{tabular}{|c|c|c|c|c|c||c|c|}
\hline
& \begin{tabular}{@{}c@{}}\textbf{Data} \\ \textbf{Augmentation}\end{tabular}& 
\begin{tabular}{@{}c@{}}\textbf{IoU} \\ \textbf{background}\end{tabular}
& \begin{tabular}{@{}c@{}}\textbf{IoU} \\ \textbf{polyp}\end{tabular}
& \begin{tabular}{@{}c@{}}\textbf{IoU} \\ \textbf{lumen}\end{tabular}
& \begin{tabular}{@{}c@{}}\textbf{IoU} \\ \textbf{spec.}\end{tabular}
& \begin{tabular}{@{}c@{}}\textbf{IoU} \\ \textbf{mean}\end{tabular}
& \begin{tabular}{@{}c@{}}\textbf{Acc} \\ \textbf{mean}\end{tabular}\\
\hline \hline
\multicolumn{8}{|l|}{\emph{FCN8 performance}} \\ \hline
4 classes & None & 86.36 & 38.51 & \textbf{43.97} & 32.98 &  50.46 &  87.40\\ 
\hline
3 classes & None & 84.66 & 47.55 & 36.93 &  --   & 56.38 & 86.08 \\ 
\hline
2 classes & None & \textbf{94.62} & 50.85 &   --   &  --   &  \textbf{72.74} & \textbf{94.91}\\ 
\hline
4 classes & Combination   & 88.81 & \textbf{51.60} & 41.21 & 38.87 & 55.13 & 89.69  \\
\hline \hline
\multicolumn{8}{|l|}{\emph{State-of-the-art methods}} \\ \hline
\cite{bernal2014polyp,bernal2014discarding,bernal2013impact}& - &73.93 & 22.13 & 23.82 & \textbf{44.86} & 41.19& 75.58  \\ 
\hline
\end{tabular}
\end{center} 
\caption{Results on the tests set: FCN8 with respect to previously published methods.}
\label{tab:results_segmentation}
\setlength\tabcolsep{6pt} 
\end{table}

The segmentation results on the test set are reported in Table \ref{tab:results_segmentation} and show a clear improvement of FCN8 over previously published methods. The following improvements can be observed when comparing previously published methods to 4-class FCN8 model trained with data augmentation: 15\% in IoU for background (mucosa), 29\% in IoU for polyps, 18\% in IoU for lumen, 14\% in mean IoU and 14\% in mean accuracy. FCN8 is still outperformed by traditional methods when it comes to specular highlights class. However, it is important to note that specular highlights class is used by hand-crafted methods to reduce false positive ratio of polyp detection and from our analysis it looks like FCN model is able to segment well polyps even when ignoring this class. For example, the best mean IoU of 72.74 \% and mean accuracy of 94.91\% are obtained by 2-class model without additional data augmentation.

\begin{figure}[t!]
\centering
\subfigure[]{\includegraphics[width=0.7\textwidth]{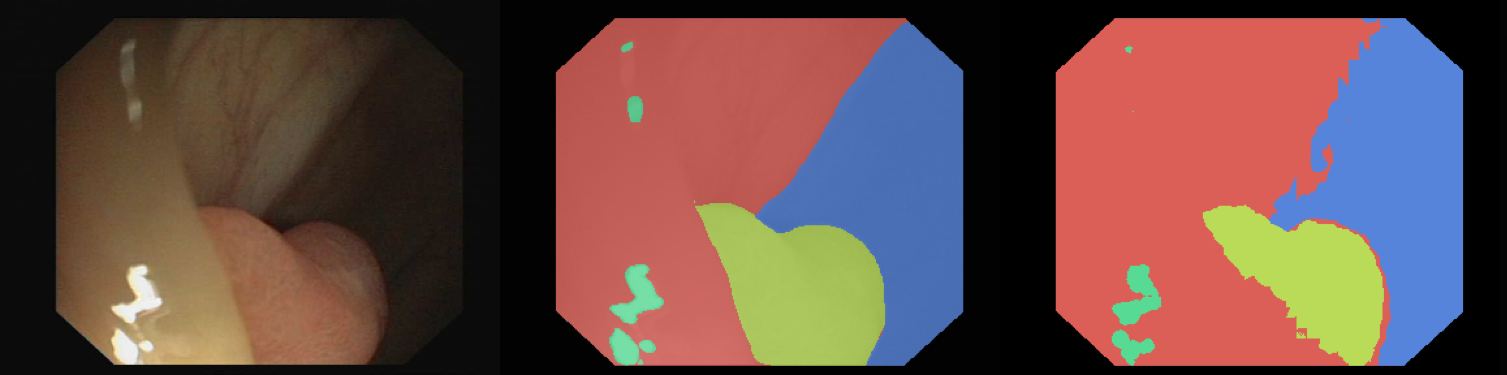}}
\subfigure[]{\includegraphics[width=0.7\textwidth]{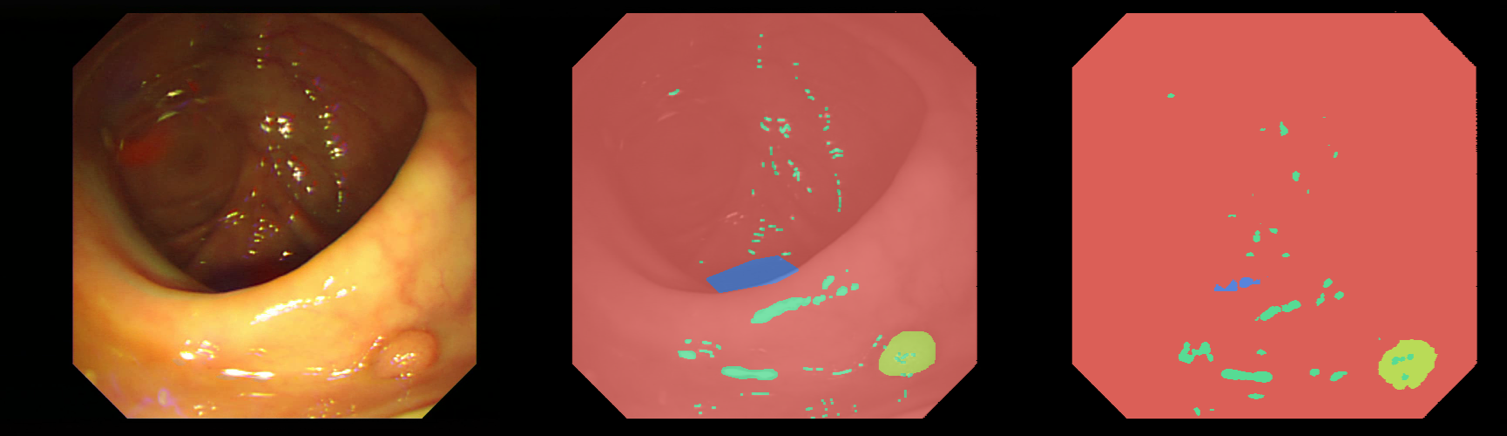}}
\subfigure[]{\includegraphics[width=0.7\textwidth]{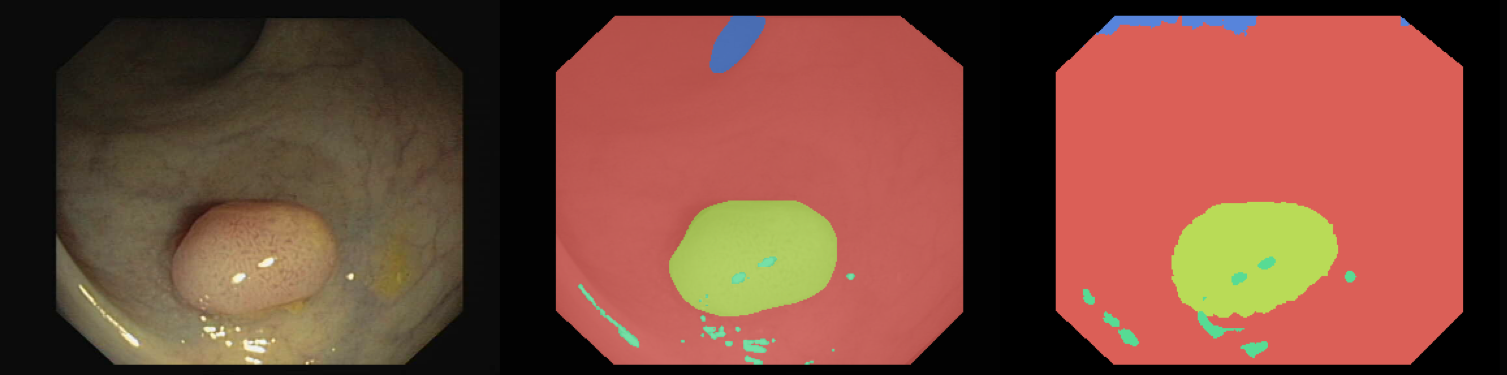}}
\subfigure[]{\includegraphics[width=0.7\textwidth]{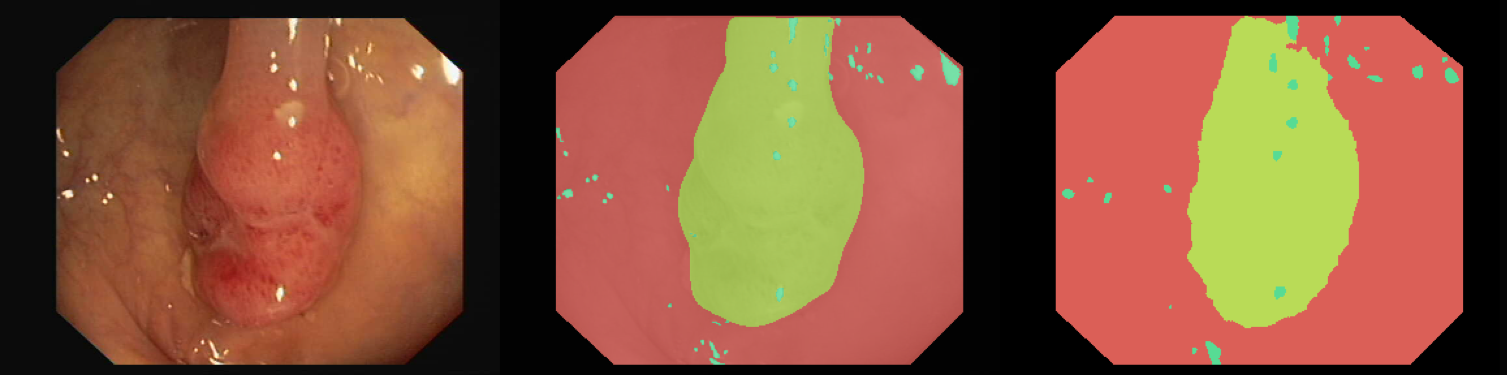}}
\subfigure[]{\includegraphics[width=0.7\textwidth]{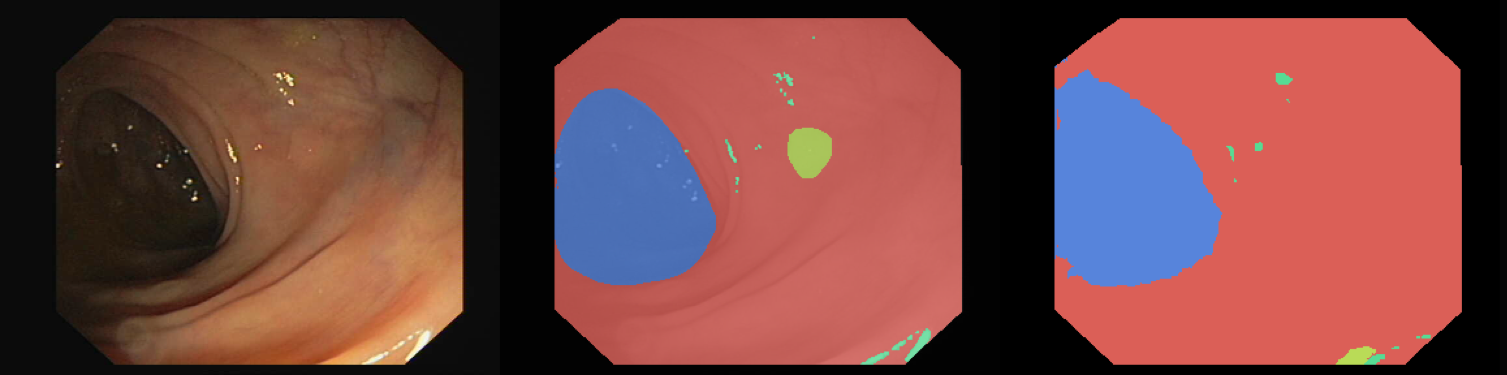}}
\subfigure[]{\includegraphics[width=0.7\textwidth]{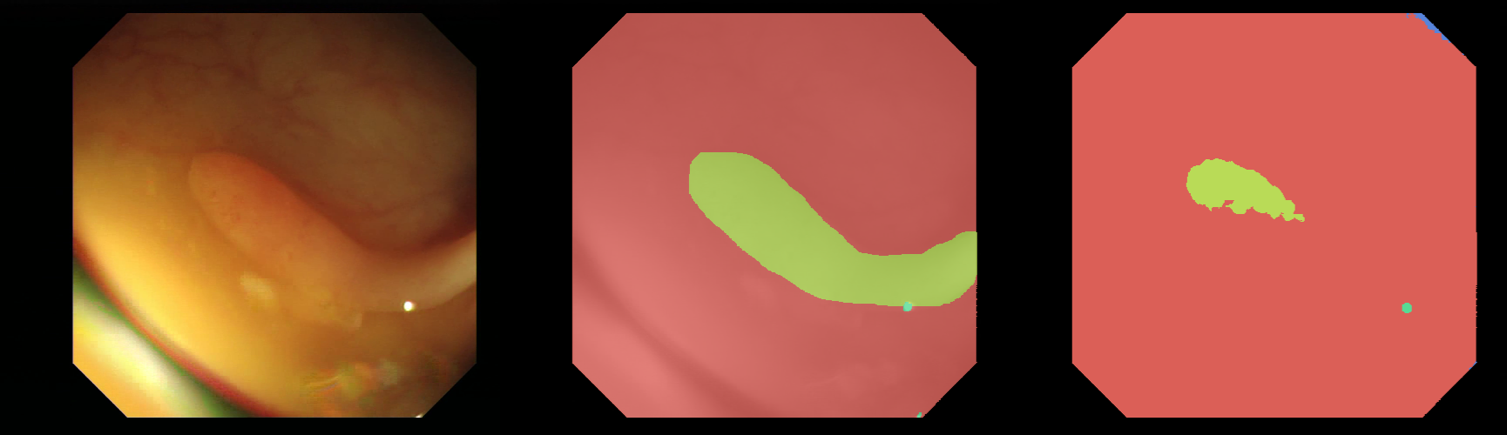}}
\caption{Examples of predictions for 4-class FCN8 model. Each sub-figure represents a single frame, a ground truth annotation and a prediction image. We use the following color coding in the annotations: red for background (mucosa), blue for lumen, yellow for polyp and green specularity. (a, b, c, d) show correct polyp segmentation, whereas (e, d) show incorrect polyp segmentation.}
\label{fig:predictions}
\end{figure}

Figure \ref{fig:predictions} shows qualitative results of 4-class FCN8 model trained with data augmentation. From left to right, each row shows a colonoscopy frame, followed by the corresponding ground truth annotation and FCN8 prediction. Rows 1 to 4 show correct segmentation masks, with very clean polyp segmentation. Rows 5 and 6 show failure modes of the model, where polyps have been missed or under-segmented. In row 5, the small polyp is missed by our segmentation methond while, in row 6, the polyp is under-segmented. All cases exhibit decent lumen segmentation and good background (mucosa) segmentation. 

\subsection{Polyp localization}

\begin{figure}[t!]
\centering
\includegraphics[width=\textwidth]{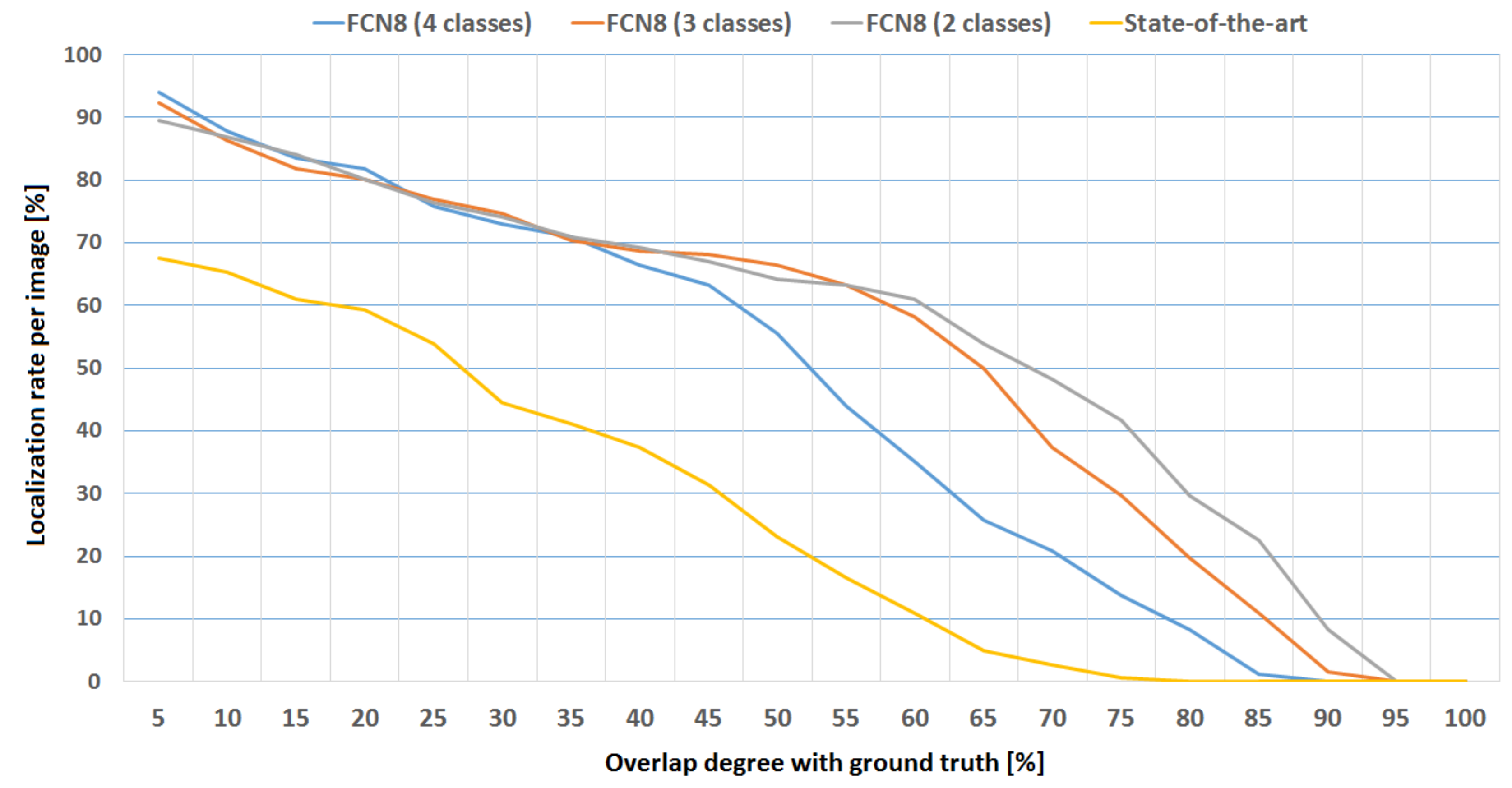}
\caption{Localization rate of polyps as a function of IoU. The x-axis represents the degree of overlap between ground truth and model prediction. The y-axis represents the percentage of correctly localized polyps. Different color plots represent different models: FCN8 with 4 classes, FCN8 with 3 classes, FCN8 with 2 classes and previously published method \cite{bernal2014polyp} (refereed to as state-of-the-art in the plot)}
\label{fig:local}
\end{figure}

Endoluminal scene segmentation can be seen as a proxy to proper polyp detection in colonoscopy video. In order to understand how well suited FCN is to localize polyps, we perform a last experiment. In this experiment, we compute the polyp localization rate as a function of IoU between the model prediction and the ground truth. We can compute this IoU per frame, since our dataset contains a maximum of one polyp per image. This analysis describes the ability of a given method to cope with polyp appearance variability and stability on polyp localization. 

The localization results are presented in Figure \ref{fig:local} and show a significant improvement when comparing FCN8 variants to the previously published method \cite{bernal2014polyp}. For example, when considering a correct polyp localization to have at least 50\% IoU, we observe an increase of 40\% in polyp localization rate. As a general trend, we observe that architectures trained using fewer number of classes achieve a higher IoU, though the polyp localization difference starts to be more visible when really high overlapping degrees are imposed. Finally, as one would expect, we observe that the architectures that show better results in polyp segmentation are the ones that show better results in polyp localization.

\section{Conclusions} 
\label{sec:conclusions}

In this paper, we have introduced an extended benchmark for endoluminal scene semantic segmentation. The benchmark includes extended annotations of polyps, background (mucosa), lumen and specular highlights. The dataset provides the standard training, validation and test splits for machine learning practitioners and will be publicly available upon paper acceptance. Moreover, standard metrics for the comparison haven been defined; with the hope to speed-up the research in the endoluminal scene segmentation area.

Together with the dataset, we provided new baselines based on fully convolutional networks, which outperformed by a large margin previously published results, without any further post-processing. We extended the proposed pipeline and used it as proxy to perform polyp detection. Due to the lack of non-polyp frames in the dataset, we reformulated the task as polyp localization. Once again, we highlighted the superiority of deep learning based models over traditional hand-crafted approaches. As expected and contrary to hand-crafted segmentation methods, when considering polyp segmentation, deep learning based approaches do not suffer from specular highlights, showing the robustness of the learnt features towards saturation zones in colonoscopy images. Moreover, given that FCN not only excels in terms of performance but also allows for nearly real-time processing, it has a great potential to be included in future DSS for colonoscopy.

Knowing the potential of deep learning techniques, efforts in the medical imaging community should be devoted to gather larger labeled datasets as well as designing deep learning architectures that would be better suited to deal with colonoscopy data. This paper pretends to make a first step towards novel and more accurate DSS by making all code and data publicly available, paving the road for more researcher to contribute to the endoluminal scene segmentation domain.

\begin{acknowledgements}
The authors would like to thank the developers of Theano \cite{Theano-2016short} and Keras \cite{chollet2015keras}. We acknowledge the support of the following agencies for research funding and computing support: Imagia Inc., Spanish government through funded project AC/DC TRA2014-57088-C2-1-R and iVENDIS (DPI2015-65286-R), SGR projects 2014-SGR-1506, 2014-SGR-1470 and 2014-SGR-135and TECNIOspring-FP7-ACCIÓ grant, by the FSEED and NVIDIA Corporation for the generous support in the form of different GPU hardware units.

\end{acknowledgements}

\bibliographystyle{spmpsci}      
\bibliography{biblio}   

\end{document}